\documentclass{article}

\usepackage{iclr2027_conference,times}

\iclrfinalcopy
\AddToHook{cmd/maketitle/after}{\lhead{Preprint.}}

\usepackage[utf8]{inputenc}
\usepackage[T1]{fontenc}
\usepackage{hyperref}
\usepackage{url}
\usepackage{booktabs}
\usepackage{graphicx}
\usepackage{amsmath}
\usepackage{amsfonts}
\usepackage{microtype}
\usepackage{multirow}
\usepackage{float}
\usepackage{tikz}
\usetikzlibrary{positioning,arrows.meta}
\usepackage{algorithm}
\usepackage{algpseudocode}

\definecolor{Gray}{gray}{0.9}
\definecolor{LightCyan}{rgb}{0.88,1,1}
\definecolor{red}{RGB}{144,0,32}
\definecolor{warmblack}{rgb}{0.0, 0.26, 0.26}
\definecolor{purple}{rgb}{0.4, 0.01, 0.24}
\definecolor{tawny}{rgb}{0.8, 0.34, 0.0}

\definecolor{blue}{RGB}{18,78,173} 

\hypersetup{
    colorlinks=true,
    citecolor=blue,
    linkcolor=blue,
    urlcolor=blue,
}

\newcommand{\osradded}[1]{{#1}}

\DeclareMathOperator*{\argmax}{arg\,max}

\title{Online Surrogate Repair: Decoupling High-Fidelity Feedback from Search Length in Closed-Loop Discovery}

\author{%
  Xiaotang Feng\textsuperscript{1,2,3}
  \quad Philip Torr\textsuperscript{1,3}
  \quad Bruno Andreis\textsuperscript{1,3}\\
  \textsuperscript{1}Department of Engineering Science, University of Oxford\\
  \textsuperscript{2}Department of Physics, University of Oxford
  \qquad
  \textsuperscript{3}Slater Labs\\
  \texttt{xiaotang.feng@stcatz.ox.ac.uk}\\
  \texttt{\{bruno.andreis,philip.torr\}@eng.ox.ac.uk}
}

\begin{document}

\maketitle

\begin{figure}[H]
  \centering
  \includegraphics[width=0.90\textwidth]{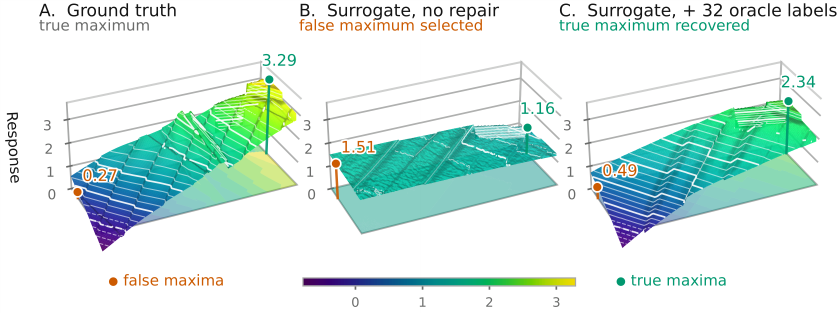}
  \caption{OSR corrects optimizer-relevant error. (A) Ground truth;
  (B) an unrepaired surrogate selects a false maximum; (C) 32
  EI-selected oracle labels restore the true maximum.}
  \label{fig:repair3d}
\end{figure}

\begin{abstract}
Closed-loop AI scientists can generate candidate designs at low marginal
computational cost, whereas reliable feedback may require wet-lab synthesis,
characterization, or high-fidelity computation. Addressing this imbalance
through custom laboratory automation remains infrastructure-intensive and
costly, while replacing new experiments with a fixed surrogate leaves
persistent model errors that can be amplified by optimization. We propose
\emph{online surrogate repair} (OSR), a closed-loop algorithm that uses sparse
high-fidelity evaluations to update the surrogate throughout a longer agent
search conducted primarily with inexpensive surrogate feedback. An acquisition
rule selects which designs from the agent's accumulated proposals receive
high-fidelity evaluation, and the resulting labels update the surrogate used
in subsequent episodes. Across controlled synthetic environments, we
demonstrate that improving global surrogate fit does not necessarily reduce
maximum regret, whereas Q90-UCB and expected improvement (EI) substantially reduce
regret by directing evaluations toward regions that determine the optimizer's
decisions. \osradded{Across LLM and Chemeleon workflows, Online EI reduces the amount of oracle queries required by 84.3--90.3\% for matching discovery
counts and by 77.9--81.6\% for matching the quality of the best material found.}
Online surrogate repair introduces a third
feedback regime between fixed-surrogate operation and high-fidelity feedback
after every episode, separating the frequency of high-fidelity evaluation from
the duration of the agent's search.
\end{abstract}

\section{Introduction}
\label{sec:intro}

AI scientists built on large language models increasingly automate hypothesis
generation, tool use, computational experimentation, and iterative planning
\citep{lu2024aiscientist,boiko2023coscientist}. Related systems now span
materials design, quantum experiments, and automated theory formation
\citep{jia2024llmatdesign,cao2024kagents,jagadish2026autocog}. Their reasoning
and proposal generation can scale at comparatively low computational cost,
whereas reliable scientific feedback may require wet-lab synthesis,
characterization, or high-fidelity computation. Discovery can therefore be
limited by validated outcomes rather than by the supply of candidate designs.

This imbalance is especially visible in computational materials discovery.
Crystal generators can propose structures at scale
\citep{XieFGBJ22,zeni2025generative,park2025chemeleon}, and learned pipelines
have expanded the number of candidate stable materials
\citep{merchant2023scaling}. Those candidates must still be ranked or validated
by interatomic potentials, electronic-structure calculations, or experiments.
Static benchmarks evaluate property prediction, stability ranking, or
one-shot generation \citep{dunn2020benchmarking,riebesell2025matbench,
betala2025lemat}, but do not reveal how approximation errors alter a policy that
repeatedly acts on model feedback. MADE instead evaluates complete discovery
policies under a sequential oracle budget \citep{made}.

Self-driving laboratories address the feedback bottleneck by automating
synthesis, handling, and characterization
\citep{burger2020robotic,kusne2020closedloop,szymanski2023alab,
rapp2024sample,sheng2024electrochemistry,seifrid2022sdl}. They reduce human
intervention and experimental latency, but require specialized infrastructure,
substantial integration effort, and continued expenditure on each physical
evaluation \citep{abolhasani2023rise,pilon2026robochem,zhang2025ivoryos}. At
the other limit, a fixed surrogate permits a long search without further
high-fidelity evaluation, but its errors persist and are sampled non-uniformly
by optimization. Maximization preferentially selects overestimated regions, so
a localized false maximum can determine the search trajectory even when global
predictive error is low
\citep{trabucco2021com,riebesell2025matbench,made}.

We introduce \emph{online surrogate repair} (OSR), a third feedback regime in
which most interactions use inexpensive surrogate feedback while sparse
high-fidelity evaluations update it during the optimization process. The AI
scientist retains control of proposal generation and design selection, while a
separate acquisition rule chooses archived proposals for oracle evaluation and
appends the returned labels to the surrogate context used in subsequent
episodes \citep{hollmann2025tabpfn,tabpfn}. Because the purpose of the
surrogate is to guide search, repair is evaluated by downstream optimization
rather than by global fit alone.

Our contributions are:

\begin{itemize}
\item We formulate online surrogate repair, in which a scientific agent
operates under inexpensive surrogate feedback while a separate acquisition
rule allocates a limited number of oracle queries to update the surrogate
during the search.
\item We show that global prediction error and maximum regret are distinct
repair objectives, and identify acquisition rules that substantially improve
the surrogate's optimization performance.
\item We demonstrate that online surrogate repair reduces the oracle feedback
required for closed-loop optimization in controlled synthetic environments
and for open-ended materials discovery on MADE under both LLM and non-LLM
proposal policies.
\end{itemize}

These results show that a limited oracle budget can improve the surrogate
throughout a closed-loop search, rather than requiring an oracle evaluation
after every agent episode.

\section{Related Work}

\subsection{Scientific agents and closed-loop discovery}

Scientific agents couple language-model planning with external tools and
feedback \citep{lu2024aiscientist,boiko2023coscientist}. In materials science,
LLMatDesign uses tool-mediated self-reflection \citep{jia2024llmatdesign};
MOFGPT and MOFGen combine language models with property predictors, generators,
and simulators \citep{badrinarayanan2025mofgpt,inizan2025system}; and LLEMA uses
LLM-guided evolutionary search \citep{abhyankar2025accelerating}. Existing benchmarks test domain reasoning and scenario- or project-level
discovery \citep{zhang2025matscibench,song2025evaluating}, while MADE focuses
on budgeted, closed-loop materials discovery \citep{made}. OSR is proposal-policy
agnostic: it changes how costly evidence updates feedback and can accompany
either an LLM agent or a fixed generator.

\subsection{Active learning, Bayesian optimization, and multi-fidelity search}

Active learning and Bayesian optimization allocate expensive evaluations with
surrogates and acquisition functions
\citep{lookman2019active,jones1998ego,garnett_bayesoptbook_2023}. Their use in
materials has motivated benchmarks of sequential and active-learning policies
\citep{rohr2020benchmarking,wang2022benchmarking} and closed-loop laboratory
demonstrations \citep{kusne2020closedloop}. In standard Bayesian optimization,
acquisition is itself the proposal policy. Multi-fidelity optimization also
chooses among information sources \citep{poloczek2017mis}, while Gemini learns
a cross-fidelity correction that drives a Bayesian optimizer
\citep{hickman2021gemini}. OSR leaves search to the scientific agent:
acquisition only chooses which archived proposal receives a high-fidelity
label to repair future feedback. Agent episodes and oracle evaluations can
therefore vary independently.

\subsection{Surrogate verifiers and tabular foundation models}

Predictive accuracy can be misaligned with discovery performance. Matbench
tests property prediction on fixed datasets \citep{dunn2020benchmarking}, while
Matbench Discovery ranks a fixed stability pool and identifies consequential
false positives near the stability boundary \citep{riebesell2025matbench}.
Learned interatomic potentials make large-scale screening practical
\citep{batatia2023foundation}, but their errors become consequential under
distribution shift. Ensemble and Bayesian interatomic potentials seek
calibrated epistemic uncertainty \citep{busk2023graph,coscia2025blips}, while
probabilistic stability models incorporate convex-hull uncertainty into active
learning \citep{novick2024probabilistic}. These methods improve or quantify the
model, while OSR asks how sparse labels should repair a surrogate already embedded in
a longer discovery loop.

Tabular foundation models permit new measurements to be incorporated as
additional context rows without task-specific training
\citep{hollmann2025tabpfn,tabpfn,tabicl}. This makes repair immediate, rather
than requiring a retraining job after every oracle query. Active in-context
learning selects labels from a fixed pool to improve global prediction
\citep{pittorino2026active}. We acquire labels from an evolving archive of
agent proposals and measure their effect on subsequent closed-loop
optimization.


\section{Methods}

Closed-loop AI scientists generate candidates far more cheaply than they obtain
reliable feedback. Fixed surrogates enable long searches but preserve
exploitable errors while high-fidelity evaluation after every episode is reliable
but costly. OSR separates search from verification: the agent searches with
inexpensive surrogate feedback, while a sparse outer loop evaluates
decision-relevant archived proposals and uses the labels to repair future
feedback. We first identify acquisition rules for optimization-critical errors,
then place the strongest in the full agent loop and evaluate OSR on synthetic and real-world benchmarks.

\subsection{Problem Definition}
\label{sec:repair}

Let $\mathcal{D}_N = \{(x_i, y_i)\}_{i=1}^{N}$ be the initial context of
expensive measurements, where $x$ is a mixed-type tabular design and
$y=f(x)$ is the oracle response for a hidden world $f$. An oracle query
invokes a physical measurement or high-fidelity computation in deployment, and
an exact or higher-fidelity simulated evaluator in our benchmarks. $Q$ is the
additional query budget beyond $\mathcal{D}_N$.

A tabular foundation model conditioned on a context
$\mathcal{D}$ returns a predictive distribution $p_{\mathcal{D}}(y \mid x)$
in a single forward pass, with mean $\mu_{\mathcal{D}}(x)$
\citep{hollmann2025tabpfn,tabpfn}. We write
$q_{\alpha,\mathcal{D}}(x)$ for the value below which this distribution
places probability $\alpha$, so $q_{0.90,\mathcal{D}}(x)$ is its
optimistic upper end. Appending the returned measurement as a labelled row to
the context is the entire model update, and no parameters are trained. Despite
low global error, a limited context can retain false maxima that an optimizer
will exploit. We define \emph{surrogate repair} as selecting designs for the $Q$ additional oracle queries and incorporating their returned labels into $\mathcal{D}$ to improve subsequent surrogate predictions.

At each repair step $t$, an acquisition rule selects a design
$x_{t+1}$ using only the current context and the surrogate's own
predictions; the oracle returns $y_{t+1} = f(x_{t+1})$, and the context
becomes $\mathcal{D}_{t+1} = \mathcal{D}_t \cup \{(x_{t+1}, y_{t+1})\}$. We
measure the repaired surrogate on two axes: global fit, as the normalized
error $E$ (NRMSE) against a hidden audit sample of the world, and
optimizer-facing quality, as the normalized maximum regret
\begin{equation}
  R_{\max}(\mathcal{D}) \;=\;
  \frac{f(x^{\star}) - f(\hat{x}_{\mathcal{D}})}{R_w},
  \qquad
  \hat{x}_{\mathcal{D}} \;=\; \argmax_{x} \mu_{\mathcal{D}}(x),
  \label{eq:rmax}
\end{equation}
where $x^{\star}$ is the best design found by a high-budget reference search on the
world itself, $\hat{x}_{\mathcal{D}}$ is the best design found by the same
search procedure on the surrogate, and $R_w$ is a robust response range
that makes regret comparable across worlds. The improvements
$\Delta R$ and $\Delta E$ report the initial value minus the final value,
so positive is better. The acquisition rules below differ only in how they
choose $x_{t+1}$, and all other components are held fixed in our experiments.

\subsection{Online Surrogate Repair for Closed-loop AI Scientists}
\label{sec:online}

\begin{figure}[t]
  \centering
  \includegraphics[width=\textwidth]{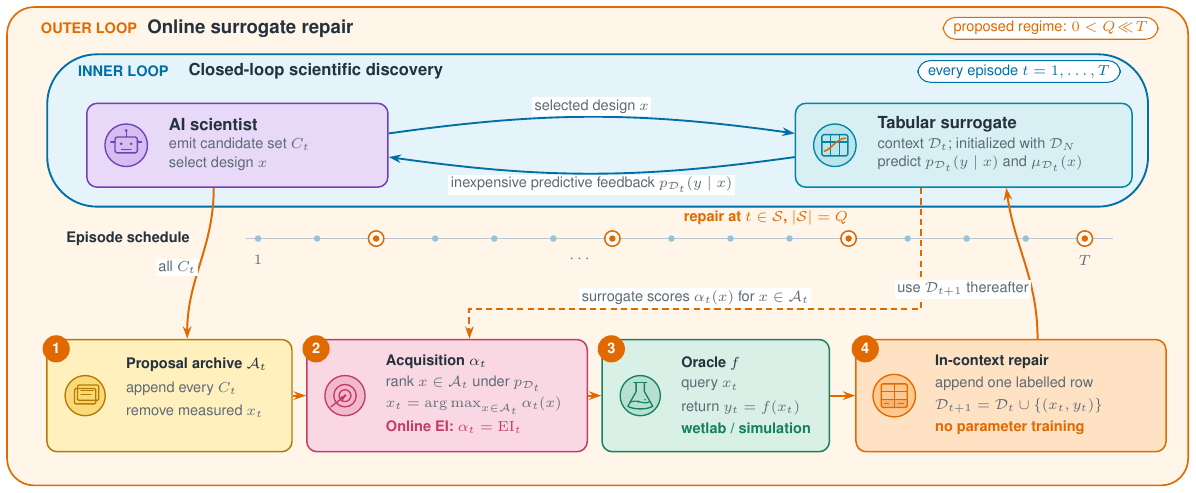}
  \caption{Online surrogate repair separates a frequent discovery loop from
  sparse surrogate updates. Candidate sets accumulate in $\mathcal A_t$; at
  $t\in\mathcal S$, acquisition selects $x_t$ for oracle evaluation, and the
  returned label is appended to $\mathcal D_t$ for subsequent episodes. Online
  EI uses $\alpha_t=\mathrm{EI}_t$.}
  \label{fig:online-repair}
\end{figure}

Across $T$ episodes, full feedback has $Q=T$, a fixed surrogate $Q=0$, and OSR
$0<Q\ll T$, with surrogate feedback each episode and $Q$ repairs scheduled
online.

Figure~\ref{fig:online-repair} illustrates the dual-loop structure, and
Algorithm~\ref{alg:online} gives the complete procedure. Every emitted candidate is recorded in an append-only archive $\mathcal{A}_t$,
and a design
leaves the archive only by being measured. The oracle is queried on a
fixed schedule $\mathcal{S} \subset \{1, \dots, T\}$ of $Q$ evenly spaced
episodes. At $t \in \mathcal{S}$ an acquisition rule $\alpha$ from
Section~\ref{sec:acquisition} selects
\begin{equation}
  x_t \;=\; \argmax_{x \in \mathcal{A}_t} \; \alpha_t(x),
  \label{eq:onlineacq}
\end{equation}
the oracle returns $y_t = f(x_t)$, and the labelled pair is appended to
the surrogate's context, so every subsequent episode is scored by the
repaired surrogate. The measured design need not be the episode's own
proposal. Instead, the label goes to whichever archived design the acquisition
method rates highest under the current surrogate. The framework accepts any
repair rule as $\alpha$. Our experiments identify expected improvement as the
strongest choice, and we call this instantiation Online EI.

\begin{algorithm}[H]
\caption{Online surrogate repair under sparse experimental feedback}
\label{alg:online}
\begin{algorithmic}[1]
\Require initial labelled context $\mathcal{D}_N$, agent, oracle $f$,
  episodes $T$, schedule $\mathcal{S}$, $|\mathcal{S}| = Q$
\State $\mathcal{D} \gets \mathcal{D}_N$
  \Comment{the surrogate starts from the initial context}
\State $\mathcal{A} \gets \emptyset$
  \Comment{the archive fills with the agent's emissions}
\For{$t = 1$ \textbf{to} $T$}
  \State agent emits candidates $C_t$ and selects $x$ using $p_{\mathcal{D}}$
  \State $\mathcal{A} \gets \mathcal{A} \cup C_t$
  \State return $p_{\mathcal{D}}(y \mid x)$ to the agent
    \Comment{surrogate feedback, every episode}
  \If{$t \in \mathcal{S}$}
    \State $x_t \gets \argmax_{x \in \mathcal{A}} \alpha_t(x)$
      \Comment{Eq.~\eqref{eq:onlineacq}}
    \State $y_t \gets f(x_t)$;\quad
      $\mathcal{D} \gets \mathcal{D} \cup \{(x_t, y_t)\}$
      \Comment{one query repairs future feedback}
    \State $\mathcal{A} \gets \mathcal{A} \setminus \{x_t\}$
  \EndIf
\EndFor
\end{algorithmic}
\end{algorithm}

A static repair variant spends the same $Q$ queries before the campaign and then
freezes the surrogate, isolating the value of allocating measurements online.

\subsection{Acquisition Rules}
\label{sec:acquisition}
Below we introduce the acquisition rules used in our surrogate repair experiments.

\subsubsection{Target agnostic methods}

These acquisition rules select labels without reference to the surrogate's estimate of
the optimum, they expand the context by improving coverage and global error.

\paragraph{Random.} Distribution-matched random sampling draws a design
from the world's own design distribution, labels it, and updates the
surrogate's context, $x_{t+1} \sim p(x)$. The surrogate plays no role in
selection, making this the control for every other method at the same query
budget.

\paragraph{Global-UQ.} Classical active learning allocates labels to reduce predictive
uncertainty globally \citep{mackay1992information,cohn1996active}. A bounded candidate set
$\mathcal{C}_t$ is constructed from the geometry of the current context
(midpoints of the longest edges of a nearest-neighbour graph over the
labelled designs), and the rule selects
\begin{equation}
  x_{t+1} \;=\; \argmax_{x \in \mathcal{C}_t}
  \bigl[\, q_{0.90,t}(x) - q_{0.10,t}(x) \,\bigr],
  \label{eq:globaluq}
\end{equation}
the design with the largest 80\% predictive interval. This is the natural
policy when the target is global fit.

\subsubsection{Optimum seeking methods}

These rules aim the query budget directly at the optimum. They share one
procedure: run a fixed surrogate-only search over the design space and query
the design that maximizes a pointwise surrogate score. They differ in how that
score reads the predictive distribution. \textsc{Peak} validates the design
the surrogate itself would select, while \textsc{Q90-UCB} validates the design
the surrogate considers most plausibly excellent.

\paragraph{Peak.} This rule naively validates the predicted optimum. It runs
the surrogate-only search to approximately maximize the predictive mean
$\mu_t(x)$, labels the incumbent $\hat{x}_{\mathcal{D}_t}$, and re-optimizes
after every update. The rule aims directly at the maximum but uses only
the surrogate's point estimate. It can therefore repair false maxima only one
at a time by spending an oracle query at each location.

\paragraph{Q90-UCB.} The same procedure read at the model's optimistic
upper quantile instead of its point prediction:
\begin{equation}
  x_{t+1} \;=\; \argmax_{x} \; q_{0.90,t}(x).
  \label{eq:q90}
\end{equation}
This is an upper-confidence rule expressed directly in the model's own
quantiles \citep{srinivas2010gpucb}. It queries designs whose plausible
upside is largest. A large
upper quantile highlights two kinds of design that oracle queries can
distinguish: designs the surrogate overestimates and designs that are genuine
optima. The label is useful in both cases,
since it collapses the quantile of an overestimated design and confirms a
genuine one, and the search then moves to the next candidate.

\paragraph{G2P($k$).} A hybrid of \textsc{Global-UQ} and
\textsc{Peak}: spend the first $k$ queries with \textsc{Global-UQ} and the
remaining $Q - k$ with \textsc{Peak}. We set $k = Q/2$, yielding G2P(16) at
$Q = 32$, so half of the budget improves global coverage before peak
validation begins.

\subsubsection{Improvement aware acquisition}

\paragraph{EI.} Expected improvement
\citep{mockus1978application,jones1998ego} references the incumbent measurement
rather than the surrogate's landscape alone. With
incumbent $y_t^{\max} = \max_{(x,y) \in \mathcal{D}_t} y$, the score of a
design is its expected margin over the best label seen so far,
\begin{equation}
  \mathrm{EI}_t(x) \;=\;
  \mathbb{E}_{p_t(y \mid x)}\!\left[\, (y - y_t^{\max})_{+} \,\right],
  \label{eq:ei}
\end{equation}
computed as the mean of $\max(q - y_t^{\max}, 0)$ over the model's full predicted distribution, so the tail that drives improvement is integrated
rather than truncated. Selection follows the same surrogate-only search
as the optimum seeking methods: at each step,
$x_{t+1} = \argmax_{x} \mathrm{EI}_t(x)$, the oracle label is appended to
the context, and the incumbent is updated before the next step.

Unlike \textsc{Q90-UCB}, \textsc{EI} integrates the full upper tail and
references the verified incumbent, so it stops prioritizing regions that
cannot improve the best observed response.


\section{Experiments}
\label{sec:experiments}

Our experiments proceed in three stages. The
static study isolates the acquisition hypothesis of Section~\ref{sec:acquisition}
on a large matrix of worlds. The synthetic world study places the repair
methods inside the full loop of Section~\ref{sec:online} and evaluates a
single-maximum regression task. The MADE experiment runs the same loop on a
multi-minima materials discovery target.

\subsection{Static surrogate repair}
\label{sec:results-static}

Each acquisition rule repairs the surrogate with $Q = 32$ oracle queries
on 40 synthetically generated tabular worlds, with two independent context
draws per world, at both context sizes $N = 100$ and $N = 1000$, and
under two surrogates, TabPFN-3 \citep{tabpfn} and TabICLv2
\citep{tabicl}. Different surrogate campaigns share their worlds, contexts
and audit samples exactly, so the models are directly comparable.
Table~\ref{tab:static} reports the final value and the improvement of
both measures of Section~\ref{sec:repair}.

\begin{table}
  \caption{Static repair after $Q{=}32$ oracle queries. The four central
  columns report reductions in maximum regret $\Delta R$ for each surrogate
  and context size, the final columns give averages of the reduction in maximum regret and the NRMSE $\Delta E$ over the four settings. Higher is better. Full $R_{\max}$ and NRMSE results are given in Appendix~\ref{app:static-full}. \textbf{Bold} best,
  \underline{underlined} second best per column.}
  \label{tab:static}
  \begin{center}
  \small
  \setlength{\tabcolsep}{4pt}
  \begin{tabular}{lcccccc}
    \toprule
    Method & TabPFN-100 & TabPFN-1000 & TabICL-100 & TabICL-1000 & Mean $\Delta R$ & Mean $\Delta E$ \\
    \midrule
    Q90-UCB   & \textbf{0.4978} & 0.2393 & \textbf{0.5909} & \textbf{0.3664} & \textbf{0.4236} & $-0.0002$ \\
    EI        & \underline{0.4516} & \textbf{0.2862} & \underline{0.5686} & \underline{0.3309} & \underline{0.4093} & 0.0011 \\
    Peak      & 0.4057 & \underline{0.2615} & 0.5593 & 0.3216 & 0.3870 & $-0.0017$ \\
    G2P(16)   & 0.3094 & 0.1672 & 0.4891 & 0.2617 & 0.3069 & 0.0013 \\
    Global-UQ & 0.0694 & $-0.0018$ & 0.1256 & 0.0593 & 0.0631 & \textbf{0.0046} \\
    Random    & 0.0565 & 0.0188 & 0.1456 & $-0.0129$ & 0.0520 & \underline{0.0045} \\
    \bottomrule
  \end{tabular}
  \end{center}
\end{table}

Across the four settings, Q90-UCB and EI achieve mean maximum-regret reductions
of $0.424$ and $0.409$, compared with $0.063$ and $0.052$ for Global-UQ and
Random. Conversely, Global-UQ and Random produce the largest mean NRMSE
improvements, while the optimizer-targeting methods yield little global error
improvement. Improving average prediction therefore does not remove the false maxima exploited during optimization. This is a crucial result because TabArena evaluates global predictive accuracy and TabPFN is trained for global prediction, not maximum regret, creating an objective mismatch with closed-loop discovery \citep{tabarena,hollmann2025tabpfn,tabpfn}. Figure~\ref{fig:repair3d}
visualizes this failure and its correction in a representative world. We carry Q90-UCB and EI into the online study and retain Peak and Global-UQ as ablations.

\subsection{Synthetic worlds}

Our main comparison uses an exact oracle $f(x)$, providing a controlled
analogue of experimental ground truth. Each of the 20 held-out worlds is a
frozen, seeded structural causal model (SCM) \citep{pearl2009causality}
inspired by TabPFN's sampled causal-graph prior \citep{hollmann2025tabpfn},
with a deterministic response and a single global maximum. These tasks
complement MADE's multi-minima discovery setting.
Appendix~\ref{app:synthetic-worlds} gives the construction and reference
protocol. A ReAct agent \citep{yao2023react} using DeepSeek-V4-Flash proposes
designs for $T=200$ episodes, and a TabPFN surrogate scores the selected
design as in Algorithm~\ref{alg:online}.

At the same episode budget, Oracle Control returns oracle feedback on the
agent-selected design each episode ($Q=T=200$). Online EI Control uses
surrogate feedback and EI-selected archive repairs every episode ($Q=T=200$).
At $Q=32$, online repair acquires labels at evenly spaced episodes, while static repair acquires them before search and then freezes the surrogate.
No repair ($Q=0$) provides the reference. Noise-free runs report normalized regret of the best oracle-verified design.

\begin{table}[t]
  \caption{Synthetic world results over 20 paired worlds under noise-free
  and noisy feedback. Values are final normalized regret with 95\% CIs
  (lower is better). Noise affects both the initial context and subsequent
  oracle observations, while regret is evaluated using the selected design's
  noise-free response. \textbf{Bold} best, \underline{underlined} second
  best within each block. The Online EI query-budget sweep is given in
  Appendix~\ref{app:extra-synth}.}
  \label{tab:synth}
  \begin{center}
  \small
  \begin{tabular}{lccc}
    \toprule
    Method & Oracle queries & Mean regret (95\% CI) & Median \\
    \midrule
    \multicolumn{4}{l}{\textit{Noise-free feedback}} \\
    \midrule
    Online EI Control ($Q{=}T{=}200$)     & 200 & \textbf{0.0278} (0.0100, 0.0499)    & \textbf{0.0014} \\
    Oracle Control ($Q{=}T{=}200$) & 200 & 0.0755 (0.0468, 0.1069) & 0.0592 \\
    Online EI ($Q{=}32$), TabPFN-3     & 32  & \underline{0.0552} (0.0298, 0.0843) & \underline{0.0393} \\
    Online Q90-UCB ($Q{=}32$), TabPFN-3 & 32 & 0.0600 (0.0338, 0.0887)             & 0.0455 \\
    Online Q90-UCB ($Q{=}32$), TabICLv2 & 32 & 0.0681 (0.0406, 0.0967)             & 0.0412 \\
    Archive-BO ($Q{=}32$), GP + EI     & 32  & 0.0807 (0.0490, 0.1144)             & 0.0592 \\
    Online Peak ($Q{=}32$), TabPFN-3   & 32  & 0.0976 (0.0702, 0.1259)             & 0.0928 \\
    Static EI ($Q{=}32$)              & 32  & 0.0984 (0.0686, 0.1291)             & 0.0983 \\
    Tab-AICL (Hybrid Method) ($Q{=}32$) & 32 & 0.1169 (0.0894, 0.1466)             & 0.0968 \\
    Static Global-UQ ($Q{=}32$)       & 32  & 0.1220 (0.0921, 0.1520)             & 0.1105 \\
    No repair                         & 0   & 0.1432 (0.1163, 0.1725)             & 0.1231 \\
    \midrule
    \multicolumn{4}{l}{\textit{Noisy feedback ($\sigma{=}0.05R_w$)}} \\
    \midrule
    Online EI Control ($Q{=}T{=}200$)     & 200 & \textbf{0.0639} (0.0384, 0.0914)    & \textbf{0.0488} \\
    Oracle Control ($Q{=}T{=}200$) & 200 & 0.1128 (0.0780, 0.1505) & 0.0898 \\
    Online EI ($Q{=}32$), TabPFN-3     & 32  & \underline{0.0905} (0.0587, 0.1252) & \underline{0.0732} \\
    Online Q90-UCB ($Q{=}32$), TabPFN-3 & 32 & 0.0981 (0.0672, 0.1292)             & 0.0892 \\
    Static EI ($Q{=}32$)              & 32  & 0.1124 (0.0811, 0.1442)             & 0.1069 \\
    No repair                         & 0   & 0.1546 (0.1264, 0.1830)             & 0.1447 \\
    \bottomrule
  \end{tabular}
  \end{center}
\end{table}

At matched $Q=32$, Online EI reduces mean regret by 44\% versus Static EI
and closes 76\% of the gap from no repair to Online EI Control ($Q=200$),
using 16\% of its queries (Table~\ref{tab:synth}). All methods use their current surrogate within episodes, Oracle Control
also appends labels to its context after each episode. Both EI settings
yield lower mean regret, consistent with more useful archive repairs.

We adapt two baselines to test whether the gains depend on the surrogate
or acquisition objective. Archive-BO retains the archive, schedule, and EI
acquisition but replaces TabPFN with a Gaussian process. Its mean regret of
$0.0807$ lies between Static EI and Online EI, showing that online repair's
benefit is not specific to in-context updating. Tab-AICL adapts active in-context learning
\citep{pittorino2026active} to sequential regression on the agent
archive (Appendix~\ref{app:tabaicl}). It reaches $0.1169$, last among the online methods and only
marginally ahead of Static Global-UQ. As in Section~\ref{sec:results-static},
selection for global coverage is less effective than targeting the regions
that determine the optimizer's decisions.

The online setting also reverses Q90-UCB's slight aggregate lead over EI
in the static study. Appendix~\ref{app:geometry} relates acquisition geometry
to final regret, where mean distances between successive acquisitions are $0.08$,
$0.36$, and $0.92$ times the initial-context distance scale for Peak, EI,
and Tab-AICL, respectively. EI lies between these extremes and achieves
the lowest regret, so broader dispersion alone does not explain performance.
This is consistent with the acquisition objectives: EI combines uncertainty
with improvement over a verified incumbent, while Peak uses point
predictions and Tab-AICL targets coverage. The budget sweep in
Appendix~\ref{app:extra-synth} shows diminishing returns when $Q=32$ is
doubled to $64$.

We add independent Gaussian noise ($\sigma=0.05R_w$) to all oracle
observations on the same 20 worlds, including the shared 1,000-row initial
context. Each run selects the measured design with the highest noisy label and regret uses its noise-free response. Online EI at $Q=32$ retains 19\% lower mean regret than Static EI and again has lower mean regret than Oracle Control. Online EI Control retains the lowest mean regret (Table~\ref{tab:synth}).

\subsection{MADE benchmark}

We further evaluate OSR on MADE \citep{made}, a closed-loop materials discovery benchmark. Each
environment specifies a chemical system and its known phases, and the agent
proposes crystal structures over an iterative campaign. \osradded{Following MADE's implementation}, the weaker MACE model \citep{batatia2023foundation} provides
surrogate feedback, while the stronger Orb-v3 model
\citep{rhodes2025orb} serves as the oracle. A novel structure within
$\tau = 0.1$\,eV/atom of the convex hull counts as a discovery. \osradded{Discovery yield covers all proposals, so we evaluate every method
post hoc with the oracle: restricting Online EI to its 32 search-time queries
would discard other proposals and cap its count at 32. Post-hoc results never
enter search or its feedback budget. Top-quality metrics instead use only search-time
oracle outcomes.}

We evaluate 30 chemical systems with three independent $T = 300$ runs per
system under two LLM orchestrators, DeepSeek-V4-Flash and GPT-5.6-Luna.
Online EI receives $Q = 32$ oracle queries on an evenly spaced
schedule. Two full feedback baselines receive an oracle query after every
episode ($Q = 300$). Vanilla follows the original MADE agent with MACE
candidate scoring, whereas Control adds the TabPFN surrogate to the same
stack.

\begin{table}[!t]
  \caption{\osradded{MADE: 30 systems $\times$ three repeats. Online EI uses
$Q{=}32$; Vanilla and Control use $Q{=}300$, with the same per-query repair for
Control and EI. top-quality metrics use only search-time oracle outcomes,
definitions are in Appendix~\ref{app:made-metrics}.
\textbf{Bold}/\underline{underlined}: highest/second reported value within each workflow.}}
  \label{tab:made}
\begin{center}
\scriptsize
\setlength{\tabcolsep}{2pt}
\renewcommand{\arraystretch}{1.25}
\resizebox{\textwidth}{!}{%
\begin{tabular}{@{}l*{12}{c}@{}}
\toprule
 & \multicolumn{4}{c}{DeepSeek-V4-Flash} & \multicolumn{4}{c}{GPT-5.6-Luna} & \multicolumn{4}{c}{Chemeleon+MLIP} \\
\cmidrule(lr){2-5}\cmidrule(lr){6-9}\cmidrule(lr){10-13}
Metric & Online EI & Vanilla & Control & Random & Online EI & Vanilla & Control & Random & Online EI & Vanilla & Control & Random \\
\midrule
\multicolumn{13}{l}{\osradded{Discovery (post-hoc)}} \\
EF@32 & \textbf{4.13(21)} & 0.882(51) & 1.00 (ref) & 0.303(55) & \textbf{5.91(29)} & \underline{1.035(72)} & 1.00 (ref) & 0.411(74) & \textbf{4.55(48)} & \underline{0.914(39)} & 1.00 (ref) & 0.138(25) \\
AF & \textbf{7.23(28)} & \underline{1.077(74)} & 1.00 (ref) & 0.363(34) & \textbf{6.36(20)} & \underline{1.280(77)} & 1.00 (ref) & 0.879(89) & \textbf{10.27(92)} & \underline{1.02(12)} & 1.00 (ref) & 0.209(23) \\
Disc.@32 & \textbf{53.9(28)} & 11.52(67) & \underline{13.07(72)} & 3.96(71) & \textbf{56.9(29)} & \underline{9.97(69)} & 9.63(67) & 3.96(70) & \textbf{130(14)} & 26.2(11) & \underline{28.68(53)} & 3.96(71) \\
Disc.@300 & --- & \underline{63.8(31)} & \textbf{65.4(34)} & 36.4(63) & --- & \textbf{86.3(49)} & \underline{84.6(47)} & 36.4(63) & --- & \underline{134(15)} & \textbf{141(16)} & 36.4(63) \\
Unmatched & \textbf{21/90} & 0/90 & --- & 0/90 & \textbf{7/90} & 0/90 & --- & 0/90 & \textbf{31/90} & \underline{1/90} & --- & 0/90 \\
AUDC@32 & \textbf{1.96(11)} & 0.401(21) & \underline{0.452(23)} & 0.121(21) & \textbf{2.07(11)} & \underline{0.291(22)} & 0.286(22) & 0.121(21) & \textbf{5.26(43)} & 0.828(36) & \underline{0.900(15)} & 0.121(21) \\
AUDC@300 & --- & \underline{0.243(12)} & \textbf{0.256(13)} & 0.120(21) & --- & \textbf{0.307(17)} & \underline{0.296(15)} & 0.120(21) & --- & \underline{0.561(48)} & \textbf{0.595(50)} & 0.120(21) \\
U. comps.@32 & \textbf{18.9(12)} & 9.50(58) & \underline{10.88(70)} & 3.91(70) & \textbf{19.0(11)} & \underline{5.52(50)} & 5.24(41) & 3.91(70) & \textbf{119(12)} & 24.9(10) & \underline{27.32(48)} & 3.91(70) \\
U. comps.@300 & --- & \underline{21.2(14)} & 20.9(15) & \textbf{35.3(62)} & --- & 20.3(14) & \underline{20.8(14)} & \textbf{35.3(61)} & --- & \underline{124(13)} & \textbf{128(14)} & 35.3(62) \\
U. SGs@32 & \textbf{22.03(77)} & 8.16(32) & \underline{8.57(38)} & 1.36(14) & \textbf{23.57(58)} & 7.19(46) & \underline{7.23(47)} & 1.36(14) & \textbf{23.1(18)} & 11.66(64) & \underline{13.46(58)} & 1.36(14) \\
U. SGs@300 & --- & \textbf{23.08(59)} & \underline{23.03(58)} & 3.61(31) & --- & \underline{27.16(94)} & \textbf{27.30(84)} & 3.61(31) & --- & \underline{23.5(17)} & \textbf{24.3(19)} & 3.61(31) \\
\midrule
\addlinespace[4pt]
\multicolumn{13}{l}{\osradded{Material quality at $Q{=}32$}} \\
\osradded{Top-1} & \osradded{\textbf{126.0(63)}} & \osradded{110.8(65)} & \osradded{\underline{114.3(60)}} & \osradded{49.6(59)} & \osradded{\textbf{133.7(64)}} & \osradded{\underline{94.7(48)}} & \osradded{93.9(50)} & \osradded{49.6(59)} & \osradded{\textbf{125.5(57)}} & \osradded{124.8(56)} & \osradded{\underline{124.9(54)}} & \osradded{49.6(59)} \\
\osradded{Top-5} & \osradded{\textbf{104.7(43)}} & \osradded{78.3(39)} & \osradded{\underline{84.1(34)}} & \osradded{24.0(39)} & \osradded{\textbf{111.9(49)}} & \osradded{\underline{67.2(39)}} & \osradded{66.3(39)} & \osradded{24.0(39)} & \osradded{\underline{106.1(38)}} & \osradded{103.0(37)} & \osradded{\textbf{106.1(36)}} & \osradded{24.0(39)} \\
\osradded{Top-5 AUC} & \osradded{\textbf{77.1(34)}} & \osradded{54.7(33)} & \osradded{\underline{59.8(30)}} & \osradded{13.3(23)} & \osradded{\textbf{87.3(41)}} & \osradded{\underline{39.2(29)}} & \osradded{36.8(28)} & \osradded{13.3(23)} & \osradded{\underline{89.1(34)}} & \osradded{85.1(37)} & \osradded{\textbf{90.2(32)}} & \osradded{13.3(23)} \\
\osradded{EF-top} & \osradded{\textbf{1.119(37)}} & \osradded{0.968(20)} & \osradded{\underline{1.00 (ref)}} & \osradded{0.447(50)} & \osradded{\textbf{1.546(119)}} & \osradded{\underline{1.082(74)}} & \osradded{1.00 (ref)} & \osradded{0.595(82)} & \osradded{\textbf{1.005(14)}} & \osradded{\underline{1.000(16)}} & \osradded{1.00 (ref)} & \osradded{0.411(49)} \\
\osradded{AF-top} & \osradded{\textbf{4.78(43)}} & \osradded{\underline{2.67(47)}} & \osradded{1.00 (ref)} & \osradded{0.48(17)} & \osradded{\textbf{5.44(42)}} & \osradded{\underline{2.21(29)}} & \osradded{1.00 (ref)} & \osradded{1.20(33)} & \osradded{\textbf{4.52(50)}} & \osradded{\underline{4.28(55)}} & \osradded{1.00 (ref)} & \osradded{0.21(11)} \\
\osradded{Unmatched} & \osradded{38/90} & \osradded{19/90} & \osradded{0/90} & \osradded{2/90} & \osradded{36/90} & \osradded{9/90} & \osradded{0/90} & \osradded{5/90} & \osradded{41/90} & \osradded{39/90} & \osradded{0/90} & \osradded{1/90} \\
\bottomrule
\end{tabular}}

\end{center}
\end{table}

\begin{table}[!ht]

\caption{Comparison of 6 methods on MADE at $T{=}100, Q{=}10$, DeepSeek-V4-Flash, 30 systems.}
\label{tab:made-policy100}
\begin{center}
\small
\setlength{\tabcolsep}{3pt}
\renewcommand{\arraystretch}{1.0}
\begin{tabular}{@{}l*{6}{c}@{}}
\toprule
Method & Top-1 & Top-5 & Top-5 AUC & EF-top & AF-top & Unmatched \\
\midrule
Online EI & \textbf{128.3(92)} & \textbf{85.1(53)} & \textbf{48.3(39)} & \textbf{1.400(114)} & \textbf{5.94(68)} & 12/30 \\
Q90-UCB & \underline{108.1(85)} & \underline{66.7(63)} & \underline{36.1(40)} & \underline{1.231(108)} & \underline{4.72(77)} & 10/30 \\
Original Control & 99.7(73) & 57.9(45) & 32.6(30) & 1.00 (ref) & 1.00 (ref) & 0/30 \\
Random archive & 79.9(71) & 33.3(45) & 18.4(26) & 0.880(79) & 2.06(55) & 2/30 \\
Fixed surrogate & 78.7(115) & 40.0(70) & 26.7(51) & 0.864(130) & 4.17(81) & 9/30 \\
Periodic proposal & 78.4(51) & 40.9(37) & 22.3(24) & 0.906(85) & 1.17(33) & 0/30 \\
\bottomrule
\end{tabular}

\end{center}
\end{table}

\begin{figure}
  \centering
  \includegraphics[width=0.99\textwidth]{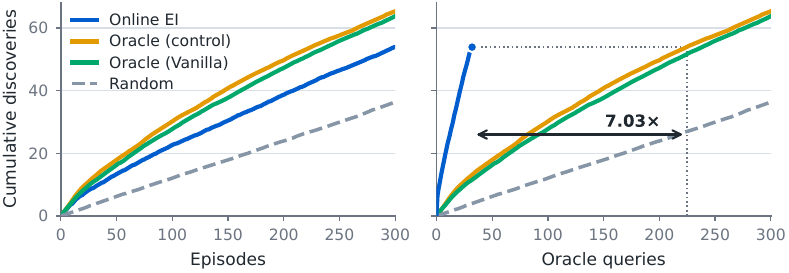}
  \caption{MADE convergence over the DeepSeek runs. Online EI ends after
  32 oracle queries. On the aggregate query curve, the full-feedback control
  required $7.03\times$ as many oracle queries to reach the same discovery
  count.}
  \label{fig:convergence}
\end{figure}

Table~\ref{tab:made} shows that Control requires 7.23(28) and 6.36(20)
times as many oracle queries to match Online EI under DeepSeek and Luna,
respectively, equivalent to 86\% and 84\% fewer queries for Online EI.
Control fails to catch up within 300 queries in 21 and 7 of the 90 runs,
so these acceleration factors are lower bounds.
Figure~\ref{fig:convergence} shows the discovery curves.

With $Q{=}32$, Online EI reaches 82\% and 67\% of the corresponding
Controls' $Q{=}300$ discovery counts and leads both full-feedback baselines
in unique compositions and space groups at matched budget.
With the non-LLM Chemeleon+MLIP workflow
\citep{park2025chemeleon,made}, it reaches 130 discoveries, 93\% of
Control's 141, with an AF of 10.27.

\osradded{To test whether these gains depend on online surrogate updates and EI acquisition, we compare six feedback policies over 100 episodes on MADE in Table~\ref{tab:made-policy100}. Alongside Online EI and the full-feedback Control, we include a fixed surrogate that disables updates, periodic validation of the current proposal, random and Q90-UCB acquisition. Online EI achieves the highest mean Top-1, Top-5, and Top-5 AUC, supporting the combined use of online surrogate repair and EI-guided query selection.}

\section{Discussions and Conclusion}

We demonstrated that global prediction error and maximum regret are misaligned surrogate objectives. In light of this we introduced \emph{online surrogate repair}, a novel algorithm
that updates the surrogate during a longer agent search with sparse oracle
queries. Controlled synthetic experiments show that global predictive accuracy
can improve without reducing optimizer-facing regret, while Online EI requires
substantially fewer oracle queries than full feedback on both synthetic worlds
and MADE. \osradded{The MADE controls further support the effectiveness of combining
online surrogate updates with EI acquisition, with Online EI achieving the
highest mean Top-1, Top-5, and Top-5 AUC among the tested policies.}

These findings suggest that feedback models should be evaluated through
the searches they guide, alongside their predictive accuracy.
\osradded{The Chemeleon results also show that the same repair procedure
extends to non-LLM proposal policies: Chemeleon produces more discoveries
than either LLM, while Luna with Online EI achieves the highest mean
Top-1 and Top-5 quality.} At matched discovery counts, Online EI
incurs modest additional LLM cost (Appendix~\ref{app:cost}), making this
approach attractive when high-fidelity evaluation is expensive relative
to continued search.

Our main implementation uses a tabular representation of the discovery
problem and a tabular foundation model that can incorporate new labels in
context. Most closed-loop discovery problems can be expressed in this form, as
our MADE experiment illustrates, but some environments may fall out-of-distribution of the model's capabilities which we did not encounter or test in our experiments.
We have not tested OSR or Online EI with agents using test-time training or
reinforcement learning, with other surrogate classes, or with more advanced and optimization-aware scheduling.

\subsection*{AI Use Statement}

In addition to the LLM-based experimental workflows described in this paper,
we used generative AI tools to polish the manuscript and correct grammatical
errors, assist with writing code, assist with creating an editable diagram
in TikZ, and support literature search and identification of related work.
The authors reviewed the AI-assisted text, tested the AI-assisted code,
and checked suggested references against the original papers.
We take responsibility for the final content of this work, including
its text, figures, code, results, and scientific claims.

\subsection*{Reproducibility Statement}

The online surrogate repair procedure is described in
Algorithm~\ref{alg:online}, with the problem formulation and acquisition
rules given in Sections~\ref{sec:repair} and~\ref{sec:acquisition}.
Section~\ref{sec:experiments} describes the experimental settings,
surrogate and oracle models, baselines, query budgets, and evaluation
protocols.
Appendix~\ref{app:synthetic-worlds} details the synthetic-world construction,
initial contexts, audit samples, and reference-search protocol.
Appendix~\ref{app:made-metrics} defines the MADE evaluation metrics.
Additional query-budget comparisons and details of the Tab-AICL baseline
adaptation are provided in Appendices~\ref{app:extra-synth}
and~\ref{app:tabaicl}, respectively. Source code is provided in the supplementary material.

\bibliographystyle{iclr2027_conference}
\bibliography{references}

@Article{abolhasani2023rise,
author="Abolhasani, Milad
and Kumacheva, Eugenia",
title="The rise of self-driving labs in chemical and materials sciences",
journal="Nature Synthesis",
year="2023",
month="Jun",
day="01",
volume="2",
number="6",
pages="483--492",
issn="2731-0582",
doi="10.1038/s44160-022-00231-0",
url="https://doi.org/10.1038/s44160-022-00231-0"
}

@Article{boiko2023coscientist,
author="Boiko, Daniil A.
and MacKnight, Robert
and Kline, Ben
and Gomes, Gabe",
title="Autonomous chemical research with large language models",
journal="Nature",
year="2023",
month="Dec",
day="01",
volume="624",
number="7992",
pages="570--578",
issn="1476-4687",
doi="10.1038/s41586-023-06792-0",
url="https://doi.org/10.1038/s41586-023-06792-0"
}

@Article{burger2020robotic,
author="Burger, Benjamin
and Maffettone, Phillip M.
and Gusev, Vladimir V.
and Aitchison, Catherine M.
and Bai, Yang
and Wang, Xiaoyan
and Li, Xiaobo
and Alston, Ben M.
and Li, Buyi
and Clowes, Rob
and Rankin, Nicola
and Harris, Brandon
and Sprick, Reiner Sebastian
and Cooper, Andrew I.",
title="A mobile robotic chemist",
journal="Nature",
year="2020",
month="Jul",
day="01",
volume="583",
number="7815",
pages="237--241",
issn="1476-4687",
doi="10.1038/s41586-020-2442-2",
url="https://doi.org/10.1038/s41586-020-2442-2"
}

@article{cao2024kagents, title={Automating quantum computing laboratory experiments with an agent-based AI framework}, volume={6}, ISSN={2666-3899}, url={http://dx.doi.org/10.1016/j.patter.2025.101372}, DOI={10.1016/j.patter.2025.101372}, number={10}, journal={Patterns}, publisher={Elsevier BV}, author={Cao, Shuxiang and Zhang, Zijian and Alghadeer, Mohammed and Fasciati, Simone D. and Piscitelli, Michele and Bakr, Mustafa and Leek, Peter and Aspuru-Guzik, Alán}, year={2025}, month=Oct, pages={101372} }

@inproceedings{tabarena,
 author = {Erickson, Nick and Purucker, Lennart and Tschalzev, Andrej and Holzm\"{u}ller, David and Desai, Prateek and Salinas, David and Hutter, Frank},
 booktitle = {Advances in Neural Information Processing Systems},
 doi = {10.52202/085713-0519},
 editor = {D. Belgrave and C. Zhang and H. Lin and R. Pascanu and P. Koniusz and M. Ghassemi and N. Chen},
 pages = {},
 publisher = {Curran Associates, Inc.},
 title = {TabArena: A Living Benchmark for Machine Learning on Tabular Data},
 url = {https://proceedings.neurips.cc/paper_files/paper/2025/file/1697e3fb412da11dc9488249f9e7bbc9-Paper-Datasets_and_Benchmarks_Track.pdf},
 volume = {38, Main Conference},
 year = {2025}
}

@misc{tabpfn,
      title={TabPFN-3: Technical Report}, 
      author={Léo Grinsztajn and Klemens Flöge and Oscar Key and Felix Birkel and Philipp Jund and Brendan Roof and Mihir Manium and Shi Bin Hoo and Magnus Bühler and Anurag Garg and Dominik Safaric and Jake Robertson and Benjamin Jäger and Simone Alessi and Adrian Hayler and Vladyslav Moroshan and Lennart Purucker and Philipp Singer and Alan Arazi and Julien Siems and Jan Hendrik Metzen and Georg Grab and Nick Erickson and Siyuan Guo and Eliott Kalfon and Simon Bing and David Salinas and Clara Cornu and Lilly Charlotte Wehrhahn and Diana Kriuchkova and Kursat Kaya and Lydia Sidhoum and Marie Salmon and Jerry Chen and Madelon Hulsebos and Yann LeCun and Samuel Müller and Bernhard Schölkopf and Sauraj Gambhir and Noah Hollmann and Frank Hutter},
      year={2026},
      eprint={2605.13986},
      archivePrefix={arXiv},
      primaryClass={cs.LG},
      url={https://arxiv.org/abs/2605.13986}, 
}

@misc{hickman2021gemini,
      title={Gemini: Dynamic Bias Correction for Autonomous Experimentation and Molecular Simulation}, 
      author={Riley J. Hickman and Florian Häse and Loïc M. Roch and Alán Aspuru-Guzik},
      year={2021},
      eprint={2103.03391},
      archivePrefix={arXiv},
      primaryClass={stat.ML},
      url={https://arxiv.org/abs/2103.03391}, 
}

@Article{hollmann2025tabpfn,
author="Hollmann, Noah
and M{\"u}ller, Samuel
and Purucker, Lennart
and Krishnakumar, Arjun
and K{\"o}rfer, Max
and Hoo, Shi Bin
and Schirrmeister, Robin Tibor
and Hutter, Frank",
title="Accurate predictions on small data with a tabular foundation model",
journal="Nature",
year="2025",
month="Jan",
day="01",
volume="637",
number="8045",
pages="319--326",
issn="1476-4687",
doi="10.1038/s41586-024-08328-6",
url="https://doi.org/10.1038/s41586-024-08328-6"
}

@misc{jagadish2026autocog,
      title={Closing the Loop to Discover Psychological Theories with an Automated Cognitive Scientist}, 
      author={Akshay K. Jagadish and Younes Strittmatter and Nori Jacoby and George Kachergis and Eric Schulz and Nathaniel Daw and Suyog H. Chandramouli and Thomas L. Griffiths},
      year={2026},
      eprint={2606.26448},
      archivePrefix={arXiv},
      primaryClass={q-bio.NC},
      url={https://arxiv.org/abs/2606.26448}, 
}

@Article{jones1998ego,
author="Jones, Donald R.
and Schonlau, Matthias
and Welch, William J.",
title="Efficient Global Optimization of Expensive Black-Box Functions",
journal="Journal of Global Optimization",
year="1998",
month="Dec",
day="01",
volume="13",
number="4",
pages="455--492",
issn="1573-2916",
doi="10.1023/A:1008306431147",
url="https://doi.org/10.1023/A:1008306431147"
}

@Article{kusne2020closedloop,
author="Kusne, A. Gilad
and Yu, Heshan
and Wu, Changming
and Zhang, Huairuo
and Hattrick-Simpers, Jason
and DeCost, Brian
and Sarker, Suchismita
and Oses, Corey
and Toher, Cormac
and Curtarolo, Stefano
and Davydov, Albert V.
and Agarwal, Ritesh
and Bendersky, Leonid A.
and Li, Mo
and Mehta, Apurva
and Takeuchi, Ichiro",
title="On-the-fly closed-loop materials discovery via Bayesian active learning",
journal="Nature Communications",
year="2020",
month="Nov",
day="24",
volume="11",
number="1",
pages="5966",
issn="2041-1723",
doi="10.1038/s41467-020-19597-w",
url="https://doi.org/10.1038/s41467-020-19597-w"
}

@Article{lookman2019active,
author="Lookman, Turab
and Balachandran, Prasanna V.
and Xue, Dezhen
and Yuan, Ruihao",
title="Active learning in materials science with emphasis on adaptive sampling using uncertainties for targeted design",
journal="npj Computational Materials",
year="2019",
month="Feb",
day="18",
volume="5",
number="1",
pages="21",
issn="2057-3960",
doi="10.1038/s41524-019-0153-8",
url="https://doi.org/10.1038/s41524-019-0153-8"
}

@misc{lu2024aiscientist,
      title={The AI Scientist: Towards Fully Automated Open-Ended Scientific Discovery}, 
      author={Chris Lu and Cong Lu and Robert Tjarko Lange and Jakob Foerster and Jeff Clune and David Ha},
      year={2024},
      eprint={2408.06292},
      archivePrefix={arXiv},
      primaryClass={cs.AI},
      url={https://arxiv.org/abs/2408.06292}, 
}

@inproceedings{
made,
title={{MADE}: Benchmark Environments for Closed-Loop Materials Discovery},
author={Shreshth A Malik and Tiarnan Doherty and Panagiotis Tigas and Muhammed Razzak and S Roberts and Aron Walsh and Yarin Gal},
booktitle={Forty-third International Conference on Machine Learning},
year={2026},
url={https://openreview.net/forum?id=nrXxVDYMMF}
}

@Article{pilon2026robochem,
author="Pilon, Simone
and Savino, Elia
and Bayley, Oliver M.
and Vanzella, Michael
and Claros, Miguel
and Siasiaridis, Petros
and Liu, Junsong
and Lukas, Florian
and Damian, Matteo
and Tseliou, Vasilis
and Intini, Niccol{\`o}
and Slattery, Aidan
and SanJos{\'e}-Orduna, Jesus
and den Hartog, Tim
and Peters, Ron A. H.
and Gargano, Andrea F. G.
and Mutti, Francesco G.
and No{\"e}l, Timothy",
title="A flexible and affordable self-driving laboratory for automated reaction optimization",
journal="Nature Synthesis",
year="2026",
month="Apr",
day="13",
issn="2731-0582",
doi="10.1038/s44160-026-01053-0",
url="https://doi.org/10.1038/s44160-026-01053-0"
}

@misc{pittorino2026active,
      title={Active In-Context Learning for Tabular Foundation Models}, 
      author={Wilailuck Treerath and Fabrizio Pittorino},
      year={2026},
      eprint={2603.27385},
      archivePrefix={arXiv},
      primaryClass={cs.LG},
      url={https://arxiv.org/abs/2603.27385}, 
}

@inproceedings{poloczek2017mis,
 author = {Poloczek, Matthias and Wang, Jialei and Frazier, Peter},
 booktitle = {Advances in Neural Information Processing Systems},
 editor = {I. Guyon and U. Von Luxburg and S. Bengio and H. Wallach and R. Fergus and S. Vishwanathan and R. Garnett},
 pages = {},
 publisher = {Curran Associates, Inc.},
 title = {Multi-Information Source Optimization},
 url = {https://proceedings.neurips.cc/paper_files/paper/2017/file/df1f1d20ee86704251795841e6a9405a-Paper.pdf},
 volume = {30},
 year = {2017}
}

@inproceedings{
tabicl,
title={Tab{ICL}v2: A Better, Faster, Scalable, and Open Tabular Foundation Model},
author={Jingang QU and David Holzm{\"u}ller and Ga{\"e}l Varoquaux and Marine Le Morvan},
booktitle={Forty-third International Conference on Machine Learning},
year={2026},
url={https://openreview.net/forum?id=SxsyLjIfWB}
}

@Article{rapp2024sample,
author="Rapp, Jacob T.
and Bremer, Bennett J.
and Romero, Philip A.",
title="Self-driving laboratories to autonomously navigate the protein fitness landscape",
journal="Nature Chemical Engineering",
year="2024",
month="Jan",
day="01",
volume="1",
number="1",
pages="97--107",
issn="2948-1198",
doi="10.1038/s44286-023-00002-4",
url="https://doi.org/10.1038/s44286-023-00002-4"
}

@misc{rhodes2025orb,
      title={Orb-v3: atomistic simulation at scale}, 
      author={Benjamin Rhodes and Sander Vandenhaute and Vaidotas Šimkus and James Gin and Jonathan Godwin and Tim Duignan and Mark Neumann},
      year={2025},
      eprint={2504.06231},
      archivePrefix={arXiv},
      primaryClass={cond-mat.mtrl-sci},
      url={https://arxiv.org/abs/2504.06231}, 
}

@Article{riebesell2025matbench,
author="Riebesell, Janosh
and Goodall, Rhys E. A.
and Benner, Philipp
and Chiang, Yuan
and Deng, Bowen
and Ceder, Gerbrand
and Asta, Mark
and Lee, Alpha A.
and Jain, Anubhav
and Persson, Kristin A.",
title="A framework to evaluate machine learning crystal stability predictions",
journal="Nature Machine Intelligence",
year="2025",
month="Jun",
day="01",
volume="7",
number="6",
pages="836--847",
issn="2522-5839",
doi="10.1038/s42256-025-01055-1",
url="https://doi.org/10.1038/s42256-025-01055-1"
}

@article{rohr2020benchmarking, title={Benchmarking the acceleration of materials discovery by sequential learning}, volume={11}, ISSN={2041-6539}, url={http://dx.doi.org/10.1039/C9SC05999G}, DOI={10.1039/c9sc05999g}, number={10}, journal={Chemical Science}, publisher={Royal Society of Chemistry (RSC)}, author={Rohr, Brian and Stein, Helge S. and Guevarra, Dan and Wang, Yu and Haber, Joel A. and Aykol, Muratahan and Suram, Santosh K. and Gregoire, John M.}, year={2020}, pages={2696–2706} }

@article{seifrid2022sdl,
    author = {Seifrid, Martin and Pollice, Robert and Aguilar-Granda, Andr{\'e}s and Morgan Chan, Zamyla and Hotta, Kazuhiro and Ser, Cher Tian and Vestfrid, Jenya and Wu, Tony C. and Aspuru-Guzik, Al{\'a}n},
    title = {Autonomous
Chemical Experiments: Challenges and Perspectives
on Establishing a Self-Driving Lab},
    journal = {Accounts of Chemical Research},
    volume = {55},
    number = {17},
    pages = {2454-2466},
    year = {2022},
    month = {08},
    issn = {0001-4842},
    doi = {10.1021/acs.accounts.2c00220},
    url = {https://doi.org/10.1021/acs.accounts.2c00220},
    eprint = {https://pubs.acs.org/achre4/article-pdf/55/17/2454/17802147/ar2c00220.pdf},
}

@Article{sheng2024electrochemistry,
author="Sheng, Hongyuan
and Sun, Jingwen
and Rodr{\'i}guez, Oliver
and Hoar, Benjamin B.
and Zhang, Weitong
and Xiang, Danlei
and Tang, Tianhua
and Hazra, Avijit
and Min, Daniel S.
and Doyle, Abigail G.
and Sigman, Matthew S.
and Costentin, Cyrille
and Gu, Quanquan
and Rodr{\'i}guez-L{\'o}pez, Joaqu{\'i}n
and Liu, Chong",
title="Autonomous closed-loop mechanistic investigation of molecular electrochemistry via automation",
journal="Nature Communications",
year="2024",
month="Mar",
day="30",
volume="15",
number="1",
pages="2781",
issn="2041-1723",
doi="10.1038/s41467-024-47210-x",
url="https://doi.org/10.1038/s41467-024-47210-x"
}

@Article{szymanski2023alab,
author="Szymanski, Nathan J.
and Rendy, Bernardus
and Fei, Yuxing
and Kumar, Rishi E.
and He, Tanjin
and Milsted, David
and McDermott, Matthew J.
and Gallant, Max
and Cubuk, Ekin Dogus
and Merchant, Amil
and Kim, Haegyeom
and Jain, Anubhav
and Bartel, Christopher J.
and Persson, Kristin
and Zeng, Yan
and Ceder, Gerbrand",
title="An autonomous laboratory for the accelerated synthesis of inorganic materials",
journal="Nature",
year="2023",
month="Dec",
day="01",
volume="624",
number="7990",
pages="86--91",
issn="1476-4687",
doi="10.1038/s41586-023-06734-w",
url="https://doi.org/10.1038/s41586-023-06734-w"
}

@InProceedings{trabucco2021com,
  title = 	 {Conservative Objective Models for Effective Offline Model-Based Optimization},
  author =       {Trabucco, Brandon and Kumar, Aviral and Geng, Xinyang and Levine, Sergey},
  booktitle = 	 {Proceedings of the 38th International Conference on Machine Learning},
  pages = 	 {10358--10368},
  year = 	 {2021},
  editor = 	 {Meila, Marina and Zhang, Tong},
  volume = 	 {139},
  series = 	 {Proceedings of Machine Learning Research},
  month = 	 {18--24 Jul},
  publisher =    {PMLR},
  url = 	 {https://proceedings.mlr.press/v139/trabucco21a.html}
}

@Article{zhang2025ivoryos,
author="Zhang, Wenyu
and Hao, Lucy
and Lai, Veronica
and Corkery, Ryan
and Jessiman, Jacob
and Zhang, Jiayu
and Liu, Junliang
and Sato, Yusuke
and Politi, Maria
and Reish, Matthew E.
and Greenwood, Rebekah
and Depner, Noah
and Min, Jiyoon
and El-khawaldeh, Rama
and Prieto, Paloma
and Trushina, Ekaterina
and Hein, Jason E.",
title="IvoryOS: an interoperable web interface for orchestrating Python-based self-driving laboratories",
journal="Nature Communications",
year="2025",
month="Jun",
day="04",
volume="16",
number="1",
pages="5182",
issn="2041-1723",
doi="10.1038/s41467-025-60514-w",
url="https://doi.org/10.1038/s41467-025-60514-w"
}

@article{cohn1996active, title={Active Learning with Statistical Models}, volume={4}, ISSN={1076-9757}, url={http://dx.doi.org/10.1613/jair.295}, DOI={10.1613/jair.295}, journal={Journal of Artificial Intelligence Research}, publisher={AI Access Foundation}, author={Cohn, D. A. and Ghahramani, Z. and Jordan, M. I.}, year={1996}, month=Mar, pages={129–145} }

@article{mackay1992information, title={Information-Based Objective Functions for Active Data Selection}, volume={4}, ISSN={1530-888X}, url={http://dx.doi.org/10.1162/neco.1992.4.4.590}, DOI={10.1162/neco.1992.4.4.590}, number={4}, journal={Neural Computation}, publisher={MIT Press - Journals}, author={MacKay, David J. C.}, year={1992}, month={July}, pages={590–604} }

@incollection{mockus1978application,
  author    = {Mockus, Jonas and Tie{\v{s}}is, Vytautas and {\v{Z}}ilinskas, Antanas},
  title     = {The Application of {Bayesian} Methods for Seeking the Extremum},
  booktitle = {Towards Global Optimisation 2},
  editor    = {Dixon, L. C. W. and Szeg{\H{o}}, G. P.},
  publisher = {North-Holland Publishing Company},
  address   = {Amsterdam},
  pages     = {117--129},
  year      = {1978},
  isbn      = {0-444-85171-2}
}

@inproceedings{srinivas2010gpucb,
author = {Srinivas, Niranjan and Krause, Andreas and Kakade, Sham and Seeger, Matthias},
title = {Gaussian process optimization in the bandit setting: no regret and experimental design},
year = {2010},
isbn = {9781605589077},
publisher = {Omnipress},
address = {Madison, WI, USA},
booktitle = {Proceedings of the 27th International Conference on International Conference on Machine Learning},
pages = {1015–1022},
numpages = {8},
location = {Haifa, Israel},
series = {ICML'10}
}

@inproceedings{
yao2023react,
title={ReAct: Synergizing Reasoning and Acting in Language Models},
author={Shunyu Yao and Jeffrey Zhao and Dian Yu and Nan Du and Izhak Shafran and Karthik R Narasimhan and Yuan Cao},
booktitle={The Eleventh International Conference on Learning Representations },
year={2023},
url={https://openreview.net/forum?id=WE_vluYUL-X}
}

@book{pearl2009causality, place={Cambridge}, edition={2}, title={Causality}, publisher={Cambridge University Press}, author={Pearl, Judea}, year={2009}}

@Article{park2025chemeleon,
author="Park, Hyunsoo
and Onwuli, Anthony
and Walsh, Aron",
title="Exploration of crystal chemical space using text-guided generative artificial intelligence",
journal="Nature Communications",
year="2025",
month="May",
day="12",
volume="16",
number="1",
pages="4379",
issn="2041-1723",
doi="10.1038/s41467-025-59636-y",
url="https://doi.org/10.1038/s41467-025-59636-y"
}

@Article{dunn2020benchmarking,
author="Dunn, Alexander
and Wang, Qi
and Ganose, Alex
and Dopp, Daniel
and Jain, Anubhav",
title="Benchmarking materials property prediction methods: the Matbench test set and Automatminer reference algorithm",
journal="npj Computational Materials",
year="2020",
month="Sep",
day="15",
volume="6",
number="1",
pages="138",
issn="2057-3960",
doi="10.1038/s41524-020-00406-3",
url="https://doi.org/10.1038/s41524-020-00406-3"
}

@Article{zeni2025generative,
author="Zeni, Claudio
and Pinsler, Robert
and Z{\"u}gner, Daniel
and Fowler, Andrew
and Horton, Matthew
and Fu, Xiang
and Wang, Zilong
and Shysheya, Aliaksandra
and Crabb{\'e}, Jonathan
and Ueda, Shoko
and Sordillo, Roberto
and Sun, Lixin
and Smith, Jake
and Nguyen, Bichlien
and Schulz, Hannes
and Lewis, Sarah
and Huang, Chin-Wei
and Lu, Ziheng
and Zhou, Yichi
and Yang, Han
and Hao, Hongxia
and Li, Jielan
and Yang, Chunlei
and Li, Wenjie
and Tomioka, Ryota
and Xie, Tian",
title="A generative model for inorganic materials design",
journal="Nature",
year="2025",
month="Mar",
day="01",
volume="639",
number="8055",
pages="624--632",
issn="1476-4687",
doi="10.1038/s41586-025-08628-5",
url="https://doi.org/10.1038/s41586-025-08628-5"
}

@Article{merchant2023scaling,
author="Merchant, Amil
and Batzner, Simon
and Schoenholz, Samuel S.
and Aykol, Muratahan
and Cheon, Gowoon
and Cubuk, Ekin Dogus",
title="Scaling deep learning for materials discovery",
journal="Nature",
year="2023",
month="Dec",
day="01",
volume="624",
number="7990",
pages="80--85",
issn="1476-4687",
doi="10.1038/s41586-023-06735-9",
url="https://doi.org/10.1038/s41586-023-06735-9"
}

@misc{betala2025lemat,
      title={LeMat-GenBench: A Unified Evaluation Framework for Crystal Generative Models},
      author={Siddharth Betala and Samuel P. Gleason and Ali Ramlaoui and Andy Xu and Georgia Channing and Daniel Levy and Clémentine Fourrier and Nikita Kazeev and Chaitanya K. Joshi and Sékou-Oumar Kaba and Félix Therrien and Alex Hernandez-Garcia and Rocío Mercado and N. M. Anoop Krishnan and Alexandre Duval},
      year={2026},
      archivePrefix={arXiv},
      primaryClass={cs.LG},
      eprint={2512.04562v2},
      url={https://arxiv.org/abs/2512.04562v2},
}

@inproceedings{XieFGBJ22,
title={Crystal Diffusion Variational Autoencoder for Periodic Material Generation},
author={Tian Xie and Xiang Fu and Octavian-Eugen Ganea and Regina Barzilay and Tommi S. Jaakkola},
booktitle={International Conference on Learning Representations},
year={2022},
url={https://openreview.net/forum?id=03RLpj-tc_}
}

@misc{jia2024llmatdesign,
      title={LLMatDesign: Autonomous Materials Discovery with Large Language Models},
      author={Shuyi Jia and Chao Zhang and Victor Fung},
      year={2024},
      eprint={2406.13163},
      archivePrefix={arXiv},
      primaryClass={cond-mat.mtrl-sci},
      url={https://arxiv.org/abs/2406.13163},
}

@article{badrinarayanan2025mofgpt,
author = {Badrinarayanan, Srivathsan and Magar, Rishikesh and Antony, Akshay and Meda, Radheesh Sharma and Barati Farimani, Amir},
title = {MOFGPT: Generative Design of Metal–Organic Frameworks using Language Models},
journal = {Journal of Chemical Information and Modeling},
volume = {65},
number = {17},
pages = {9049-9060},
year = {2025},
doi = {10.1021/acs.jcim.5c01625},
note = {PMID: 40875331},
URL = {https://doi.org/10.1021/acs.jcim.5c01625},
eprint = {https://doi.org/10.1021/acs.jcim.5c01625}
}

@misc{inizan2025system,
      title={System of Agentic AI for the Discovery of Metal-Organic Frameworks},
      author={Theo Jaffrelot Inizan and Sherry Yang and Aaron Kaplan and Yen-hsu Lin and Jian Yin and Saber Mirzaei and Mona Abdelgaid and Ali H. Alawadhi and KwangHwan Cho and Zhiling Zheng and Ekin Dogus Cubuk and Christian Borgs and Jennifer T. Chayes and Kristin A. Persson and Omar M. Yaghi},
      year={2025},
      eprint={2504.14110},
      archivePrefix={arXiv},
      primaryClass={cond-mat.mtrl-sci},
      url={https://arxiv.org/abs/2504.14110},
}

@inproceedings{abhyankar2025accelerating,
title={{LLEMA}: Evolutionary Search with {LLM}s for Multi-Objective Materials Discovery},
author={Nikhil Abhyankar and Sanchit Kabra and Saaketh Desai and Chandan K. Reddy},
booktitle={The Fourteenth International Conference on Learning Representations},
year={2026},
url={https://openreview.net/forum?id=TIqzhBvCNB}
}

@misc{song2025evaluating,
      title={Evaluating Large Language Models in Scientific Discovery},
      author={Zhangde Song and Jieyu Lu and Yuanqi Du and Botao Yu and Thomas M. Pruyn and Yue Huang and Kehan Guo and Xiuzhe Luo and Yuanhao Qu and Yi Qu and Yinkai Wang and Haorui Wang and Jeff Guo and Jingru Gan and Parshin Shojaee and Di Luo and Andres M Bran and Gen Li and Qiyuan Zhao and Shao-Xiong Lennon Luo and Yuxuan Zhang and Xiang Zou and Wanru Zhao and Yifan F. Zhang and Wucheng Zhang and Shunan Zheng and Saiyang Zhang and Sartaaj Takrim Khan and Mahyar Rajabi-Kochi and Samantha Paradi-Maropakis and Tony Baltoiu and Fengyu Xie and Tianyang Chen and Kexin Huang and Weiliang Luo and Meijing Fang and Xin Yang and Lixue Cheng and Jiajun He and Soha Hassoun and Xiangliang Zhang and Wei Wang and Chandan K. Reddy and Chao Zhang and Zhiling Zheng and Mengdi Wang and Le Cong and Carla P. Gomes and Chang-Yu Hsieh and Aditya Nandy and Philippe Schwaller and Heather J. Kulik and Haojun Jia and Huan Sun and Seyed Mohamad Moosavi and Chenru Duan},
      year={2026},
      archivePrefix={arXiv},
      primaryClass={cs.AI},
      eprint={2512.15567v2},
      url={https://arxiv.org/abs/2512.15567v2},
}

@inproceedings{zhang2025matscibench,
author = {Zhang, Junkai and Gan, Jingru and Wang, Xiaoxuan and Jia, Zian and Gu, Changquan and Chen, Jianpeng and Zhu, Yanqiao and Ma, Mingyu Derek and Zhou, Dawei and Li, Ling and Wang, Wei},
title = {MatSciBench: Benchmarking the Reasoning Ability of Large Language Models in Materials Science},
year = {2026},
isbn = {9798400722592},
publisher = {Association for Computing Machinery},
address = {New York, NY, USA},
url = {https://doi.org/10.1145/3770855.3818888},
doi = {10.1145/3770855.3818888},
booktitle = {Proceedings of the 32nd ACM SIGKDD Conference on Knowledge Discovery and Data Mining V.2},
pages = {12868–12879},
numpages = {12},
location = {Republic of Korea},
series = {KDD '26}
}

@book{garnett_bayesoptbook_2023,
place={Cambridge},
title={Bayesian Optimization},
publisher={Cambridge University Press},
author={Garnett, Roman},
year={2023}
}

@article{wang2022benchmarking,
    author = {Wang, Alex and Liang, Haotong and McDannald, Austin and Takeuchi, Ichiro and Kusne, Aaron Gilad},
    title = {Benchmarking active learning strategies for materials optimization and discovery},
    journal = {Oxford Open Materials Science},
    volume = {2},
    number = {1},
    pages = {itac006},
    year = {2022},
    month = {01},
    issn = {2633-6979},
    doi = {10.1093/oxfmat/itac006},
    url = {https://doi.org/10.1093/oxfmat/itac006},
    eprint = {https://academic.oup.com/ooms/article-pdf/2/1/itac006/45033697/itac006.pdf},
}

@article{novick2024probabilistic,
    author = {Novick, Andrew and Cai, Diana and Nguyen, Quan and Garnett, Roman and Adams, Ryan and Toberer, Eric},
    title = {Probabilistic prediction of material stability: integrating convex hulls into active learning},
    journal = {Materials Horizons},
    volume = {11},
    number = {21},
    pages = {5381-5393},
    year = {2024},
    month = {11},
    issn = {2051-6347},
    doi = {10.1039/d4mh00432a},
    url = {https://doi.org/10.1039/d4mh00432a},
    eprint = {https://pubs.rsc.org/mh/article-pdf/11/21/5381/9808528/d4mh00432a.pdf},
}

@article{batatia2023foundation,
    author = {Batatia, Ilyes and Benner, Philipp and Chiang, Yuan and Elena, Alin M. and Kovács, Dávid P. and Riebesell, Janosh and Advincula, Xavier R. and Asta, Mark and Avaylon, Matthew and Baldwin, William J. and Berger, Fabian and Bernstein, Noam and Bhowmik, Arghya and Bigi, Filippo and Blau, Samuel M. and Cărare, Vlad and Ceriotti, Michele and Chong, Sanggyu and Darby, James P. and De, Sandip and Della Pia, Flaviano and Deringer, Volker L. and Elijošius, Rokas and El-Machachi, Zakariya and Fako, Edvin and Falcioni, Fabio and Ferrari, Andrea C. and Gardner, John L. A. and Gawkowski, Mikołaj J. and Genreith-Schriever, Annalena and George, Janine and Goodall, Rhys E. A. and Grandel, Jonas and Grey, Clare P. and Grigorev, Petr and Han, Shuang and Handley, Will and Heenen, Hendrik H. and Hermansson, Kersti and Ho, Cheuk Hin and Hofmann, Stephan and Holm, Christian and Jaafar, Jad and Jakob, Konstantin S. and Jung, Hyunwook and Kapil, Venkat and Kaplan, Aaron D. and Karimitari, Nima and Kermode, James R. and Kourtis, Panagiotis and Kroupa, Namu and Kullgren, Jolla and Kuner, Matthew C. and Kuryla, Domantas and Liepuoniute, Guoda and Lin, Chen and Margraf, Johannes T. and Magdău, Ioan-Bogdan and Michaelides, Angelos and Moore, J. Harry and Naik, Aakash A. and Niblett, Samuel P. and Norwood, Sam Walton and O’Neill, Niamh and Ortner, Christoph and Persson, Kristin A. and Reuter, Karsten and Rosen, Andrew S. and Rosset, Louise A. M. and Schaaf, Lars L. and Schran, Christoph and Shi, Benjamin X. and Sivonxay, Eric and Stenczel, Tamás K. and Sutton, Christopher and Svahn, Viktor and Swinburne, Thomas D. and Tilly, Jules and van der Oord, Cas and Vargas, Santiago and Varga-Umbrich, Eszter and Vegge, Tejs and Vondrák, Martin and Wang, Yangshuai and Witt, William C. and Wolf, Thomas and Zills, Fabian and Csányi, Gábor},
    title = {A foundation model for atomistic materials chemistry},
    journal = {The Journal of Chemical Physics},
    volume = {163},
    number = {18},
    pages = {184110},
    year = {2025},
    month = {11},
    issn = {0021-9606},
    doi = {10.1063/5.0297006},
    url = {https://doi.org/10.1063/5.0297006},
    eprint = {https://pubs.aip.org/aip/jcp/article-pdf/doi/10.1063/5.0297006/20801246/184110_1_5.0297006.pdf},
}

@article{busk2023graph,
    author = {Busk, Jonas and Schmidt, Mikkel N. and Winther, Ole and Vegge, Tejs and Jørgensen, Peter Bjørn},
    title = {Graph neural network interatomic potential ensembles with calibrated aleatoric and epistemic uncertainty on energy and forces},
    journal = {Physical Chemistry Chemical Physics},
    volume = {25},
    number = {37},
    pages = {25828-25837},
    year = {2023},
    month = {10},
    issn = {1463-9076},
    doi = {10.1039/d3cp02143b},
    url = {https://doi.org/10.1039/d3cp02143b},
    eprint = {https://pubs.rsc.org/cp/article-pdf/25/37/25828/8980537/d3cp02143b.pdf},
}

@inproceedings{coscia2025blips,
title={{BLIP}s: Bayesian Learned Interatomic Potentials},
author={Dario Coscia and Pim De Haan and Max Welling},
booktitle={Forty-third International Conference on Machine Learning},
year={2026},
url={https://openreview.net/forum?id=ZSZW2diTs5}
}

\clearpage
\appendix
\raggedbottom

\section{MADE evaluation metrics}
\label{app:made-metrics}

We retain MADE's discovery and diversity metrics \citep{made}. Let
$D_{\pi}(q)$ be the cumulative number of discoveries made by policy $\pi$
after $q$ oracle queries. Higher is better for every metric in
Table~\ref{tab:made}.

\begin{itemize}
  \item \textbf{Discoveries (Disc.).} The number of proposed structures that
  are stable within the specified hull threshold, absent from the initial
  reference set, and distinct from earlier proposals under MADE's structure
  matching criterion.
  \item \textbf{Area under the discovery curve (AUDC).}
  $\mathrm{AUDC}_{\pi}(q)=\frac{2}{q^2}\int_0^q D_{\pi}(u)\,du$, which
  rewards discoveries made earlier in the query sequence.
  \item \textbf{Unique compositions (U. comps.).} The number of distinct
  chemical compositions among the discovered structures.
  \item \textbf{Unique space groups (U. SGs).} The number of distinct crystal
  space groups among the discovered structures.
  \item \textbf{Enhancement factor (EF).} At a matched query budget $q$,
  $\mathrm{EF}_{\pi}(q)=D_{\pi}(q)/D_{\mathrm{Control}}(q)$. We report
  EF at $q=32$.
  \item \textbf{Acceleration factor (AF).} For each paired run, let $j$
  and $k$ be the query counts at which policy $\pi$ and Control first
  reach $D_{\pi}(32)$, respectively. We report the mean of $k/j$ over
  runs with $D_{\pi}(32)>0$. If Control does not reach the target within
  300 queries, we set $k=300$, giving a lower bound.
  \item \textbf{Unmatched.} The number of runs (out of 90) in which
  Control does not reach policy $\pi$'s discovery count at $q=32$
  within its full 300-query budget, i.e.,
  $D_{\mathrm{Control}}(300) < D_{\pi}(32)$.
  \item \osradded{\textbf{Top-1.} The highest score
  $s(x)=\max(0,100-h_0(x))$ among novel, distinct structures
  evaluated by the oracle during search, where $h_0(x)$ is the
  signed Orb-v3 energy gap to the initial convex hull in meV/atom.}

  \item \osradded{\textbf{Top-5.} The mean of the five highest
  scores, with missing entries set to zero.}

  \item \osradded{\textbf{Top-5 AUC.}
  $\frac{1}{Q}\int_0^Q \mathrm{Top5}(q)\,dq$,
  evaluated using the trapezoidal rule.}

  \item \osradded{\textbf{EF-top.} Top-1 relative to Control
  at the same oracle-query budget.}

  \item \osradded{\textbf{AF-top.} $K/Q$, where $K$ is the
  number of oracle queries Control requires to reach the
  method's Top-1 at budget $Q$.}

  \item \osradded{\textbf{Unmatched (top-quality).} The number of runs
  whose Top-1 exceeds Control's best within its full query budget.}
\end{itemize}

Under MADE's original full-feedback setting, AUDC is bounded by one because at
most one discovery is made per oracle query. Online surrogate repair can
produce multiple discoveries per oracle query, so AUDC values above one are
valid here. Except for Unmatched, \osradded{the discovery entries in}
Table~\ref{tab:made} are means over 90 runs, with parentheses giving the
standard error in the final displayed digit or digits.

\section{Synthetic world construction}
\label{app:synthetic-worlds}

Each world is one frozen SCM with a mixed-type tabular design space and a
deterministic scalar response. Following the sampled-DAG construction of
TabPFN's synthetic prior \citep{hollmann2025tabpfn}, its seed fixes the schema,
root distributions, graph, and every mechanism parameter. Separate row-sampling
streams ensure that a design has the same response whether evaluated alone or
in a batch.

A world has 14--22 observable roots drawn from six types: continuous, integer,
ordinal, nominal, Boolean, and cyclic. Continuous roots follow rescaled Beta
laws; integers are uniform over finite ranges; ordinal and nominal
probabilities are drawn from Dirichlet distributions; Booleans follow Bernoulli
laws; and cyclic values are uniform over periods of 12, 24, or 360. Roots are
encoded to $[-1,1]$ before graph evaluation, with cyclic features represented
by the sine of their phase.

Above the roots is a 5--8-layer DAG with 44--72 hidden nodes and 2--6 parents
per node. Its twelve nonlinear mechanisms are affine, neural, tree, quadratic,
Fourier, oscillator, radial basis, threshold, plateau, product, gated, and
categorical. The scalar response combines 4--7 final-layer branches, two
pairwise products, and a direct root term together with its sine. Five
mechanism profiles produce smooth multiscale, tree/threshold/plateau,
oscillatory, localized-basin, and categorical/quantized families. We use the
third disjoint cohort, W5241--W5260, balanced at four worlds per family; these
worlds are not used to train the surrogates.

For each world, preparation draws a 1,000-row initial context and a held-out
50,000-row audit sample. A guided world-level search uses 10 restarts, 20
iterations, and 500 proposals per iteration. The reference response is the
larger maximum found by the audit sample and guided search, and $R_w$ is the
audit sample's $q_{0.95}-q_{0.05}$ range. Static EI and Global-UQ select their 32 queries as a single batch from a
shared, locked pool of 10,000 first-occurrence rows using the initial surrogate. This well-established SCM construction provides a
reproducible, controlled basis for comparing acquisition rules and schedules
on identical worlds.

\section{Query-budget sweep and selection geometry}

\subsection{Online EI query-budget sweep}
\label{app:extra-synth}

\begin{table}[H]
  \caption{Online EI query-budget sweep on the 20 synthetic worlds, all with
  TabPFN. Values are final normalized regret with 95\% CIs (lower is
  better).}
  \label{tab:qsweep}
  \begin{center}
  \small
  \begin{tabular}{lccc}
    \toprule
    Method & Oracle queries & Mean regret (95\% CI) & Median \\
    \midrule
    Online EI ($Q{=}64$) & 64 & \textbf{0.0481} (0.0233, 0.0760)    & \textbf{0.0139} \\
    Online EI ($Q{=}32$) & 32 & \underline{0.0552} (0.0298, 0.0843) & \underline{0.0393} \\
    Online EI ($Q{=}16$) & 16 & 0.0775 (0.0502, 0.1060)             & 0.0813 \\
    Online EI ($Q{=}8$)  & 8  & 0.0918 (0.0599, 0.1251)             & 0.0831 \\
    \bottomrule
  \end{tabular}
  \end{center}
\end{table}

Online EI improves monotonically over the eightfold range from $Q{=}8$ to
$Q{=}64$, so its advantage does not depend on a narrow choice of budget.
Returns diminish: the step from $Q{=}8$ to $Q{=}32$ reduces mean regret by
$0.0366$, whereas doubling again to $Q{=}64$ reduces it by a further
$0.0071$. Even at $Q{=}8$, four times fewer queries than the main setting,
Online EI reaches $0.0918$, still ahead of Static EI at $Q{=}32$
($0.0984$) and well ahead of no repair ($0.1432$).

\subsection{Selection geometry of online acquisition rules}
\label{app:geometry}

We characterize the 32 acquisitions made by each online method in each world.
Designs use the world-specific $[-1,1]$ representation, with wraparound
distance for cyclic features. Let $d_0$ denote the mean pairwise RMS distance
among the 1{,}000 initial context designs. Step is the distance between
successive acquisitions, while Spread is the mean pairwise distance among all
acquired designs, and both are normalized by $d_0$. An incumbent update occurs
when an acquired label exceeds the previous verified incumbent. The Step
statistics comprise 620 successive acquisition pairs per method across the 20
worlds.

\begin{table}[H]
  \caption{Selection geometry of the four TabPFN online methods. Step and
  Spread are normalized by the initial-context scale $d_0$; incumbent updates
  are reported per world.}
  \label{tab:geometry}
  \begin{center}
  \small
  \begin{tabular}{lcccc}
    \toprule
    Method & Step / $d_0$ & Spread / $d_0$ & Incumbent updates & Mean regret \\
    \midrule
    Online Peak    & 0.079 & 0.189 & 6.1 & 0.0976 \\
    Online Q90-UCB & 0.189 & 0.255 & 6.0 & 0.0600 \\
    Online EI      & 0.356 & 0.448 & 5.6 & 0.0552 \\
    Tab-AICL       & 0.915 & 0.916 & 1.1 & 0.1169 \\
    \bottomrule
  \end{tabular}
  \end{center}
\end{table}

Step and Spread have the same ordering: Peak, Q90-UCB, EI, and Tab-AICL.
EI moves $1.9\times$ farther per step than Q90-UCB and $4.5\times$ farther
than Peak. Regret, however, is not monotonic in either measurement. Peak
concentrates its acquisitions within a small region, whereas Tab-AICL covers
nearly the full initial-context scale; their mean regrets are $0.0976$ and
$0.1169$. EI lies between these extremes and obtains the lowest mean regret,
$0.0552$. Neither local concentration nor broad coverage is therefore
sufficient for low regret. This distinction is consistent with the acquisition
objectives: Peak uses an incumbent reference without uncertainty, Tab-AICL
targets coverage without an incumbent reference, and EI combines uncertainty
with improvement over a verified incumbent.

Incumbent update frequency also does not explain the performance ordering.
Peak, Q90-UCB, and EI update the incumbent 6.1, 6.0, and 5.6 times per world,
despite their different final regrets. The three methods also take smaller
steps following an update: $0.032$ rather than $0.081$ for Peak, $0.119$
rather than $0.183$ for Q90-UCB, and $0.186$ rather than $0.350$ for EI.
EI's advantage is therefore not associated with greater dispersion immediately
after finding a new incumbent.

These measurements establish an association between acquisition geometry and
final regret, rather than a causal mechanism. Archive-BO is omitted here to compare acquisition rules under the same TabPFN surrogate.

\section{Full static surrogate repair results}
\label{app:static-full}

Table~\ref{tab:static-full} reports the complete final maximum-regret and
NRMSE matrix summarized in Table~\ref{tab:static}.

\begin{table}[H]
  \caption{Static surrogate repair with $Q{=}32$ across TabPFN-3 and
  TabICLv2 at context sizes $N\in\{100,1000\}$. \textbf{Bold} best,
  \underline{underlined} second best per metric within each block.}
  \label{tab:static-full}
  \begin{center}
  \small
  \begin{tabular}{lcccc}
    \toprule
    Method & Final $R_{\max}$ $\downarrow$ & $\Delta R$ $\uparrow$ & Final NRMSE $\downarrow$ & $\Delta E$ $\uparrow$ \\
    \midrule
    \multicolumn{5}{l}{\emph{TabPFN-3, $N{=}100$}} \\
    Q90-UCB   & \textbf{0.5672} & \textbf{0.4978} & 0.2248 & 0.0023 \\
    EI        & \underline{0.6133} & \underline{0.4516} & 0.2219 & 0.0052 \\
    Peak      & 0.6593 & 0.4057 & 0.2273 & $-0.0002$ \\
    G2P(16)   & 0.7555 & 0.3094 & 0.2227 & 0.0044 \\
    Global-UQ & 0.9955 & 0.0694 & \textbf{0.2183} & \textbf{0.0088} \\
    Random    & 1.0084 & 0.0565 & \underline{0.2197} & \underline{0.0074} \\
    \midrule
    \multicolumn{5}{l}{\emph{TabPFN-3, $N{=}1000$}} \\
    Q90-UCB   & 0.5583 & 0.2393 & 0.1736 & $-0.0006$ \\
    EI        & \textbf{0.5113} & \textbf{0.2862} & 0.1731 & $-0.0002$ \\
    Peak      & \underline{0.5361} & \underline{0.2615} & 0.1741 & $-0.0011$ \\
    G2P(16)   & 0.6303 & 0.1672 & 0.1730 & $-0.0001$ \\
    Global-UQ & 0.7993 & $-0.0018$ & \textbf{0.1723} & \textbf{0.0007} \\
    Random    & 0.7787 & 0.0188 & \underline{0.1724} & \underline{0.0005} \\
    \midrule
    \multicolumn{5}{l}{\emph{TabICLv2, $N{=}100$}} \\
    Q90-UCB   & \textbf{0.5694} & \textbf{0.5909} & 0.2313 & $-0.0009$ \\
    EI        & \underline{0.5918} & \underline{0.5686} & 0.2301 & 0.0004 \\
    Peak      & 0.6010 & 0.5593 & 0.2330 & $-0.0025$ \\
    G2P(16)   & 0.6712 & 0.4891 & 0.2284 & 0.0021 \\
    Global-UQ & 1.0347 & 0.1256 & \underline{0.2223} & \underline{0.0082} \\
    Random    & 1.0147 & 0.1456 & \textbf{0.2209} & \textbf{0.0095} \\
    \midrule
    \multicolumn{5}{l}{\emph{TabICLv2, $N{=}1000$}} \\
    Q90-UCB   & \textbf{0.4752} & \textbf{0.3664} & 0.1792 & $-0.0017$ \\
    EI        & \underline{0.5107} & \underline{0.3309} & \underline{0.1786} & $-0.0011$ \\
    Peak      & 0.5200 & 0.3216 & 0.1804 & $-0.0030$ \\
    G2P(16)   & 0.5799 & 0.2617 & \underline{0.1786} & $-0.0012$ \\
    Global-UQ & 0.7823 & 0.0593 & \textbf{0.1769} & \underline{0.0005} \\
    Random    & 0.8545 & $-0.0129$ & \textbf{0.1769} & \textbf{0.0006} \\
    \bottomrule
  \end{tabular}
  \end{center}
\end{table}

\section{Cost estimation}
\label{app:cost}

All experiments ran on two H100 SXM GPUs, with runtime dominated by
waiting for LLM responses and comparable GPU costs for Online EI and
Control. LLM charges were estimated from records collected through
OpenAI's native Responses API and OpenRouter's Chat Completions API,
with uncertainty from stochastic context caching and provider routing.
At matched mean discovery counts, Online EI spends an additional \$0.33
(DeepSeek) and \$0.68 (Luna) per trajectory relative to Control
(Table~\ref{tab:cost}), a small increase compared with a typical wet-lab
synthesis and characterization cycle, as motivated in Section~\ref{sec:intro}.

\begin{table}[H]
  \caption{Estimated costs per MADE trajectory. LLM costs are in USD;
  H100-s denotes GPU-seconds. Discovery counts follow
  Table~\ref{tab:made}. Catch-up denotes Control reaching Online EI's
  final discovery count. Cost per discovery includes LLM charges only.}
  \label{tab:cost}
  \begin{center}
  \begin{tabular}{@{}lrrrr@{}}
    \toprule
    Operating point & LLM cost (\$) & H100-s & Discoveries & \$/discovery \\
    \midrule
    \multicolumn{5}{l}{\emph{DeepSeek-V4-Flash}} \\
    Online EI, full run ($Q{=}32$)
      & 1.35 & 922 & 53.9 & 0.025 \\
    Control, catch-up (ep.~224.5)
      & 1.02 & 813 & 53.9 & 0.019 \\
    Control, matched ($Q{=}32$)
      & 0.12 & 102 & 13.07 & 0.009 \\
    Control, full run ($Q{=}300$)
      & 1.44 & 1,125 & 65.4 & 0.022 \\
    \midrule
    \multicolumn{5}{l}{\emph{GPT-5.6-Luna}} \\
    Online EI, full run ($Q{=}32$)
      & 2.02 & 342 & 56.9 & 0.036 \\
    Control, catch-up (ep.~190.6)
      & 1.34 & 387 & 56.9 & 0.024 \\
    Control, matched ($Q{=}32$)
      & 0.16 & 63 & 9.63 & 0.017 \\
    Control, full run ($Q{=}300$)
      & 2.22 & 652 & 84.6 & 0.026 \\
    \bottomrule
  \end{tabular}
  \end{center}
\end{table}

\section{Tab-AICL adaptation}
\label{app:tabaicl}

We adapt Tab-AICL's hybrid rule \citep{pittorino2026active} to
regression with one label per acquisition. Classification entropy is
replaced by the surrogate's native 80\% predictive interval width,
$q_{0.90,t}(x)-q_{0.10,t}(x)$.
We retain the original shortlist size
$N_{\mathrm{cand}}=\min(M,\max(2,\lfloor M/2\rfloor))$,
where $M=|\mathcal{A}_t|$, keeping the most uncertain archived designs.
For $B=1$, we replace batch $k$-means with the paper's Coreset criterion:
select the shortlisted design maximizing
$D_{\min,t}(x)=\min_{(x',y')\in\mathcal{D}_t}
\|z_t(x)-z_t(x')\|_2$.

The preprocessing map $z_t$ standardizes numerical features and
ordinal-encodes categorical features using the current labelled
context. The surrogate, $Q=32$ budget, query
schedule, and candidate archive match Online EI, only the selection
rule differs.

\end{document}